\documentclass[sigconf]{acmart}
\AtBeginDocument{%
  }

\copyrightyear{2026}
\acmYear{2026}
\setcopyright{cc}
\setcctype{by}
\acmConference[ISWC '26]{Proceedings of the 2026 ACM International Symposium on Wearable Computers}{October 11--15, 2026}{Shanghai, China}
\acmBooktitle{Proceedings of the 2026 ACM International Symposium on Wearable Computers (ISWC '26), October 11--15, 2026, Shanghai, China}
\acmDOI{10.1145/3830727.3834826}
\acmISBN{979-8-4007-2872-3/2026/10}

\usepackage{amsmath}
\usepackage{balance}       
\usepackage{graphics}      
\usepackage[T1]{fontenc}   
\usepackage{lipsum}
\usepackage{color}
\usepackage{colortbl}
\usepackage{booktabs}
\usepackage{textcomp}
\usepackage{microtype}        
\usepackage{ccicons}          
\usepackage{multirow}  
\usepackage{color,soul}  
\usepackage{subcaption}  
\usepackage{listings}  
\usepackage{todonotes}
\usepackage{multirow}

\usepackage[export]{adjustbox} 

\usepackage{array}
\newcolumntype{P}[1]{>{\centering\arraybackslash}p{#1}}
\newcolumntype{L}[1]{>{\raggedright\arraybackslash}p{#1}}

\usepackage[capitalize]{cleveref}

\usepackage{enumitem}

\usepackage{hyperref}

\usepackage{balance}

\begin{document}

\hypersetup{
  colorlinks=true,
  linkcolor=blue,
  urlcolor=blue
}

\title{Models Got Talent: Identifying High Performing Wearable Human Activity Recognition Models Without Training}

\author{Richard Goldman}
\affiliation{%
  \institution{Georgia Institute of Technology}
  \city{Atlanta}
  \country{USA}
}
\author{Varun Komperla}
\affiliation{%
  \institution{Georgia Institute of Technology}
  \city{Atlanta}
  \country{USA}
}
\author{Thomas Ploetz}
\affiliation{%
  \institution{Georgia Institute of Technology}
  \city{Atlanta}
  \country{USA}
}

\author{Harish Haresamudram}
\affiliation{%
  \institution{University of Illinois Urbana-Champaign}
  \city{Urbana}
  \country{USA}}



\renewcommand{\shortauthors}{Richard Goldman, Varun Komperla, Thomas Ploetz, \& Harish Haresamudram}

\begin{abstract}
    Discovering high performing model architectures for wearables-based Human Activity Recognition (HAR) applications is challenging.
    The astonishing diversity and variability due to differing sensor locations, recording apparatus, activities, etc., can cause established architectures to perform worse on datasets/tasks they were not designed for. 
    A promising complement to Neural Architecture Search (NAS) involves the development of \textit{Zero Cost Proxies (ZCPs)}, which correlate well with trained performance, but can be computed through a single forward/backward pass on a randomly sampled batch of data.
    In this paper, we investigate the effectiveness of eight ZCPs on six benchmark HAR datasets, and demonstrate that the top-predicted architectures obtain performance within 7\% of that attained by full-scale training of 2,000 randomly sampled architectures.
    Furthermore, training the top-10 predicted architectures results in performance within 2\% of full-scale training, leading to  substantial computational savings. 
    Our experiments introduce ZCPs to sensor-based HAR and demonstrate their suitability as an addition to NAS pipelines in practical scenarios.
\end{abstract}




\maketitle

\section{Introduction}
Selecting effective architectures for sensor-based Human Activity Recognition (HAR) is challenging because wearable datasets differ substantially in sensor placement, hardware, activities, users, and preprocessing \cite{stisen2015smart}. Consequently, architectures that perform well on one dataset may not transfer reliably to another. 
Neural Architecture Search (NAS) addresses this problem, but  conventionally requires training many candidate models \cite{yu2019evaluating,bender2018understanding, liu2018darts}.
Grid and random search are simpler options, with random search being highly effective while requiring substantially lower computational cost. 

Alternatively, developing \textit{proxy scores for network performance} has shown great promise.
These scores use minimal data, e.g., a single minibatch of data and a single forward/backward propagation pass, and are expected to be strongly correlated with performance obtained after completing training \cite{abdelfattah2021zero}. 
\textit{If effective}, these scores are valuable for identifying high-performing network architectures for newly collected datasets (and more broadly, other wearable sensing applications), without fully-fledged modeling efforts.
They can be used independently or to accelerate NAS pipelines \cite{abdelfattah2021zero}.

Consequently, we evaluate ZCP-guided architecture selection across \textit{six HAR datasets and 2,000 convolutional, recurrent, and Transformer} architecture candidates. 
We find that training only the highest-ranked ZCP candidate leads to performance  within 5\% of the best-performing network identified after full-scale training. 
Training the top-10 predicted models further reduces the gap to 2\%, demonstrating the ZCPs' capability for identifying strong HAR architectures without training. 
In summary, the contributions of this work are:
\textit{(i)} We present the \textit{first comprehensive study} of ZCPs for sensor-based HAR, evaluating \textit{eight proxies} across six benchmark datasets and 2,000 architectures; and
\textit{(ii) } We define suitable metrics, such as ``Talent Rate'', addressing  key deficiencies in existing metrics for analyzing the efficacy of ZCPs, paving the way for their integration and use in existing sensor pipelines. 


\section{Related Work}
\label{sec:related}

A characteristic feature of wearable applications, including HAR, is wide variety in data collection protocols, human demographics, activities, and hardware, all of which result in substantial diversity in the data collected \cite{stisen2015smart}. 
Consequently, developing suitable classifiers is a challenging endeavor, 
as established architectures can perform poorly in diverging conditions.
While grid search is effective, random search is more computationally efficient \cite{yu2019evaluating, li2020random} while retaining performance.
Neural Architecture Search (NAS) \cite{zoph2016neural} is highly effective but accompanied by \textit{substantial computational costs} \cite{li2020random}.
It has been explored for HAR through Q-learning \cite{pellatt2021fast} and evolutionary algorithms \cite{wang2021harnas, ismail2023auto}, optimizing for accuracy, latency, model size, and computational cost \cite{pellatt2021fast,wang2021harnas,nas_meets_pruning,ismail2023auto,liberis2021munas,king2025micronas}.

\citet{mellor_jacob},  first coined the term ``Zero Cost Proxy'' (ZCP), which are scores that require no (or minimal) upfront training to make predictions about performance obtained after full training. 
\citet{krishnakumar2022nasbenchsuitezeroacceleratingresearchzero} evaluate 13 such proxies, and \cite{lukasik2023evaluationzerocostproxies} provides an excellent overview, categorizing them as: Jacobian-based, pruning-based, baseline, piecewise-linear, and Hessian-based, while primarily studying ZCPs for computer vision tasks.
In contrast, our goal is to systematically evaluate the potential and efficacy of ZCPs for predicting high-performing architectures in the HAR domain. 

\vspace{-0.25 em}
\section{Evaluating Zero Cost Proxies for HAR} \label{sec:method}

Our experimental workflow comprises four steps: 

\noindent \textbf{1. Sampling Random Architectures: }
We randomly sample from a broad landscape of CNN, RNN, and Transformer architectures, varying properties such as depth, hidden/channel dimensions, kernel size, attention heads, and dropout.
These ranges span ${\approx}9.23{\times}10^{25}$ CNNs, ${\approx}9.88{\times}10^{9}$ RNNs, and ${\approx}2.33{\times}10^{4}$ Transformers (the full search-space configuration will be detailed in our code repository).
From all possible candidates, we sample 2,000 architectures per dataset (1,200 CNNs, 500 Transformers, and 300 RNNs).

\vspace{0.5em}
\noindent \textbf{2. Zero-Cost Proxies (ZCPs): }
They assign each architecture a scalar score using a single randomly sampled training batch and one forward/backward pass at initialization, avoiding the cost of full training. 
We evaluate eight ZCPs spanning gradient-, pruning-, and initialization-based criteria, seven of which follow the work of \citet{abdelfattah2021zero}. 
\texttt{jacob\_cov} measures correlations among input-output Jacobians for a given batch as an indicator of representational diversity, while \texttt{grad\_norm} uses gradient magnitude to estimate trainability. 
The pruning-based proxies \texttt{snip}, \texttt{grasp}, \texttt{synflow}, and \texttt{fisher} aggregate parameter-saliency scores based on first-order loss sensitivity, gradient-flow preservation, data-free synaptic flow, and Fisher information, respectively. 
We additionally consider two baseline metrics: \texttt{plain}, which aggregates unweighted parameter gradients, and \texttt{initial\_val\_f1}, which we introduce as performance measured before training. 
Since the scores have different scales, we convert each proxy's score into an architecture ranking and form an \texttt{ensemble} by averaging across ranks. 
These scores are inexpensive heuristics rather than precise performance predictors and can be used to steer a broader NAS procedure.

\vspace{0.5em}
\noindent \textbf{3. Full-Scale Model Training: }
\label{sec:full_scale_training}
After computing ZCPs, we fully train all sampled architectures to obtain their 
performance ranking.
This is used to quantify the computational savings from replacing exhaustive training with ZCP-guided selection.
All models are initialized using Xavier normal \cite{glorot2010understanding}, and trained for 50 epochs using Adam \cite{kingma2014adam}.
We avoid per-dataset tuning and use the same learning rate (lr=$10^{-4}$) and batch size (bs=256) for all datasets, following \cite{haresamudram2022assessing}.
The learning rate is decayed by a factor of 0.8 every 10 epochs.  

\begin{figure*}[!t]
    \centering
    \includegraphics[width=0.85\linewidth]{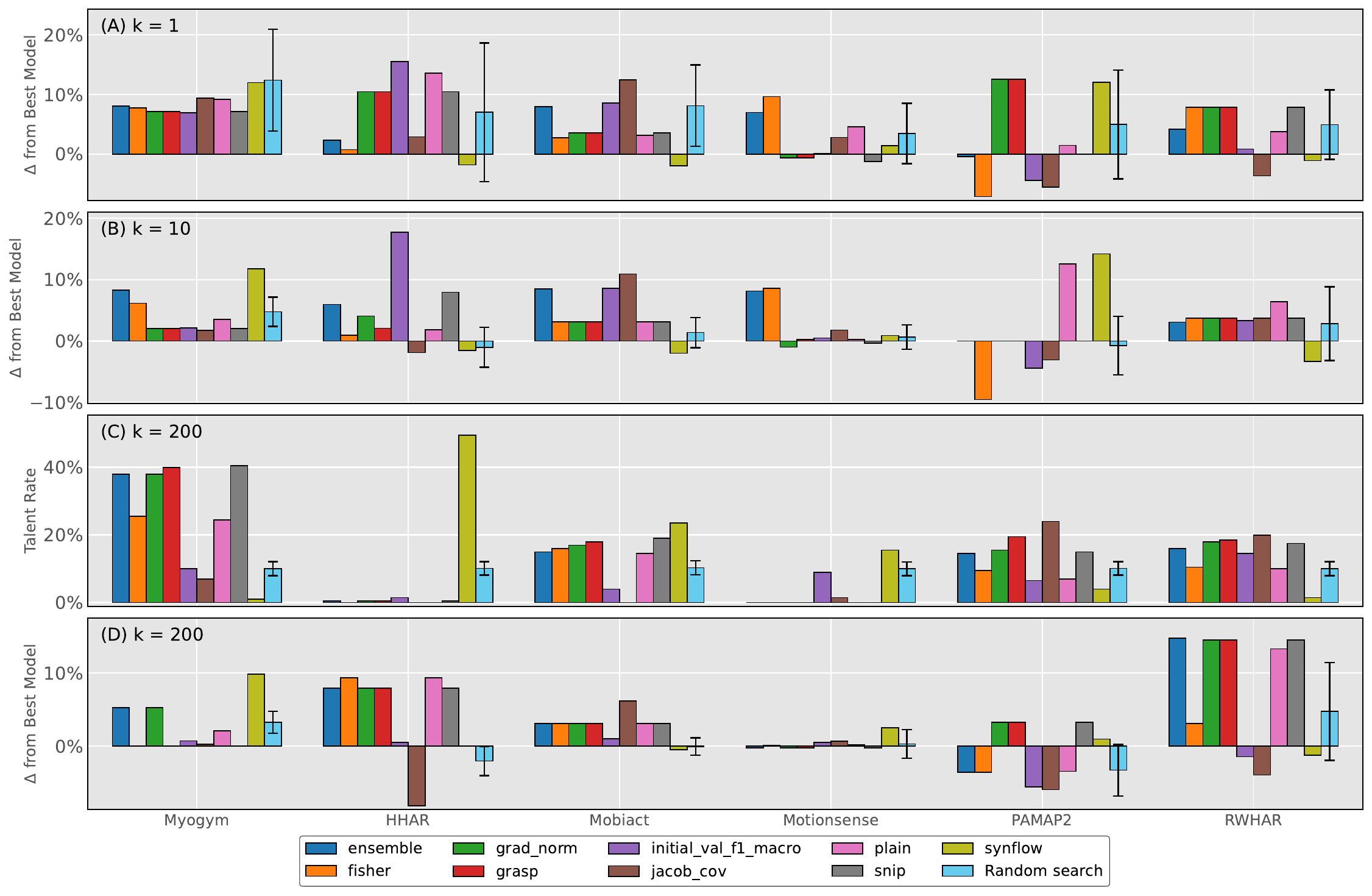}
    \caption{
    (A, B): Visualizing the best-performing proxy for $\Delta_1$ and $\Delta_{10}$ -- the performance gap to full-scale grid search is less than 7\% and 2\% respectively when training only top-1 and top-10 predicted networks respectively;
    (C, D): Talent rate -- approx.\ 20\% of the top ZCP-predicted models appear within the best fully-trained models. 
    }
    \label{fig:delta_k1}
\end{figure*}

\vspace{0.5em}
\noindent \textbf{4. Performance Evaluation:}
\label{sec:eval_metrics}
We evaluate the utility of ZCPs using the following metrics:
\begin{enumerate}[leftmargin=*, topsep=0pt, itemsep=0pt]
    \item \textbf{Performance gap, $\Delta_k$:} 
    which is the difference in performance  
    between the best architecture identified through full training, and the architecture selected using ZCPs. 
    The ZCP-selected model is chosen by first taking the top-$k$ ranked architectures, training only those $k$ candidates, and then selecting the one with the highest validation score. 
    The closer $\Delta_k$ is to zero, the better is the performance of the ZCP, as it would mean that the best-performing architectures can be identified at close to zero (or minimal) cost.
    $\Delta_1$ corresponds to the most restrictive setting, where only the top-ranked ZCP architecture is trained.
    \item \textbf{Talent Rate:} 
    We define it as the percentage of top-10\% ZCP-ranked architectures that also fall in the top 10\% after full training. 
    It focuses on whether ZCPs predict the strongest candidates.
\end{enumerate}

\vspace{-0.25 em}
\section{Experimental Settings}
\label{sec:settings}

\textit{Datasets:  }
We evaluate on \textit{six} datasets in total, covering a broad spectrum of activities, sensing locations, and number of participants: 
(i) HHAR \cite{stisen2015smart}; (ii) Mobiact \cite{chatzaki2016human}; (iii) Motionsense \cite{malekzadeh2018protecting}; (iv) Myogym  \cite{koskimaki2017myogym}; (v) PAMAP2 \cite{reiss2012introducing}; and (vi) RealWorld \cite{sztyler2016body}. 
We utilize \textit{only accelerometer data}.
For HHAR, we utilize smartwatch data.
We use five out of seven available locations for RWHAR, as data were missing from two locations for the Climbing Up and Jumping activities. 
We use data from head, chest, upper arm, waist, and shin. 

The datasets broadly contain locomotion-style activities, e.g., sitting, walking, along with biking (HHAR), jumping (RealWorld, PAMAP2), transitions (Mobiact), and 
daily living activities such as vacuum cleaning and ironing (PAMAP2).
Myogym is highly imbalanced (null class >75\%) and comprises fine-grained gym activities.

\textit{Data Pre-Processing:  }
All datasets are partitioned into training, validation, and test sets based on user IDs.
Users are randomly assigned to the three splits in a 60\%/20\%/20\% ratio, respectively.
Data are downsampled to 50 Hz to match the lowest frequency across datasets. 
Within each split, data are concatenated and  segmentation is applied with a window size of two seconds and an overlap of 50\% between segments.
We will publicly release  code at: \textcolor{blue}{\href{https://github.com/rgold5950/models-got-talent-iswc}{link}}.

\vspace{-0.25 em}
\section{Results}
\label{sec:results}

\subsection{Performance Gap  ($\Delta_1$ and $\Delta_{10}$)}
\label{sec:delta_k1}

In \cref{fig:delta_k1} (A), we visualize $\Delta_1$ for each dataset and ZCP.
For most datasets, the \textit{best-performing ZCPs have $\Delta_1$<5\%}, indicating their \textit{practical utility} in the wearables domain, as newly collected datasets or novel tasks can be effectively tackled using ZCPs.
For PAMAP2, \texttt{ensemble}, \texttt{fisher}, \texttt{jacob\_cov}, and \texttt{initial\_val\_f1\_macro}
have $\Delta_1$<0\%.
As such, these ZCPs produce higher test set performance than full-scale training with val set based model selection.

While individual ZCPs such as \texttt{fisher}, \texttt{initial\_val\_f1\_score}, and \texttt{grad\_norm} are broadly effective, the \texttt{ensemble} is the \textit{most practically relevant setting} because it does not require prior knowledge of which proxy will work best. 
By aggregating across all ZCPs, the \texttt{ensemble} reduces reliance on any single measure whose effectiveness may vary across datasets or architecture families (see \cref{sec:method} for details). 
Its comparable or stronger performance suggests that ZCP-guided search can be applied robustly in new HAR settings where the optimal proxy is unknown.
This is precisely the regime in which ZCPs are most valuable, since researchers and practitioners tackling novel sensing tasks rarely have sufficient labeled data or compute to empirically identify the best proxy or architectures.

In \cref{fig:delta_k1} (B), we extend the evaluation to top-10 predicted architectures, ranking them by their validation F1-scores.
We then quantify the gap between the overall best model and the best model identified within this top-10 subset, which we denote by $\Delta_{10}$.
We observe that \textit{the best ZCPs have $\Delta_{10} < 0\%$} for all datasets except Myogym which has \textit{$\Delta_{10}<2\%$}, demonstrating that training just the top-10 predicted architectures is sufficient to match the performance obtained after full-scale training of 2,000 networks.

Across 1,000 simulated draws, random search sometimes outperforms ZCP selection but has substantial variance.
ZCP avoids this variability and reduces risk of unfavorable random draws.
Training the top-10 ZCP-ranked candidates instead of all 2,000 reduces the number of fully trained models by 200-fold while remaining within 2\% of  exhaustive training.
This comparison illustrates how savings scale with the number of candidates, and improves with larger workflows. Furthermore, ZCPs can help prioritize training candidates after domain-informed narrowing or within NAS.

\vspace{-0.25 em}
\subsection{Talent Rates of ZCPs}
Talent rate 
indicates whether training a small ZCP-selected subset is likely to lead to strong performance (see 
\cref{fig:delta_k1} (C)).
Four out of six datasets have talent rates >$15\%$ with the highest, Myogym, at 38\%.
\texttt{grasp}, \texttt{snip}, and \texttt{grad\_norm} perform better than  \texttt{ensemble} on these datasets, indicating the effectiveness of gradient-based proxies. 
Interestingly, \texttt{synflow} is the only ZCP with a high talent rate for HHAR, highlighting the benefit of using data-free proxies. 

For Myogym, Mobiact, PAMAP2, and RWHAR, the ZCP talent rate is comparable to or better than random search.
This demonstrates that ZCPs can enrich the selected subset with high-performing architectures.
We fully train the top-10\% of ZCP-predicted models and denote the difference in performance to the top model as $\Delta_{10\%}$.
We observe that ensemble has a $\Delta_{10\%} < \Delta_{1}$ in four out of six datasets, as the larger architecture space leads to higher performance.

Top predicted architecture requires on the order of $10^{10}$ FLOPs for training, while the top-$10$ and top-$10\%$ require around $10^{11}$ and $10^{12}$ FLOPs respectively, compared to $10^{13}$ FLOPs for training all 2,000 candidates.
Random search can also be effective, but \textit{has substantial variance}, making it unreliable under fixed training budgets.
As such, the primary benefit of utilizing ZCPs  relative to random search is the reduced risk of sampling low performing architectures.

\vspace{-0.25 em}
\section{Conclusion}
In this paper, we introduced Zero-Cost Proxies (ZCPs) to wearable HAR and evaluated their potential for identifying effective classifier architectures. 
We compared the performance gap of the best ZCP-predicted architecture(s) against those identified after full-scale training. 
We observed that the gap is surprisingly low, at around 5\% when the top predicted model is trained, and around 1-2\% when the top-10 models are trained. 
Subsequently, we investigated the talent rate, i.e., the overlap between high-performing models identified through ZCPs and via full training. 
This work paves the way for the use and integration of ZCPs into HAR-based NAS pipelines, leading to more efficient discovery of high-performing architectures.

\begin{acks}
This work was partially supported by NSF IIS-2112633.
\end{acks}

\bibliographystyle{ACM-Reference-Format}
\balance
\bibliography{refs}

\end{document}